\documentclass[runningheads]{llncs}
\usepackage{times}
\usepackage{epsfig}
\usepackage{graphicx}
\usepackage{amsmath}
\usepackage{amssymb}
\usepackage{color}
\usepackage{epstopdf}
\usepackage{pdfpages}
\usepackage{subfig}
\usepackage{algorithm}
\usepackage{algpseudocode}
\usepackage{times}
\usepackage{url}
\usepackage{enumitem}
\usepackage{booktabs}
\usepackage{multirow}
\usepackage{verbatim}
\usepackage{authblk}
\usepackage{dutchcal}
\usepackage{multirow}
\usepackage{wrapfig}

\newcommand{\etal}{\textit{et~al}. }

\newcommand{\conv}[1]{$\left[\begin{array}{ll} \text{1}\times \text{1} \text{ conv}\\ \text{3}\times \text{3} \text{ conv} \end{array}\right] \times \text{#1}$}
\newcommand{\cross}[1]{#1 $\times$ #1}

\begin{document}
\title{Composition Loss for Counting, Density Map Estimation and Localization in Dense Crowds}

\titlerunning{Composition Loss for Counting in Dense Crowds}
\authorrunning{H. Idrees \etal}

\author{Haroon~Idrees\inst{1} \and Muhmmad~Tayyab\inst{5} \and Kishan~Athrey\inst{5} \and Dong~Zhang\inst{2} \and Somaya~Al-Maadeed\inst{3} \and Nasir~Rajpoot\inst{4} \and Mubarak~Shah\inst{5}}
\institute{Robotics Institute, Carnegie Mellon University \and
NVIDIA Inc. \and
Computer Science Department, Faculty of Engineering, Qatar University \and
Department of Computer Science, University of Warwick, UK \and
Center for Research in Computer Vision, University of Central Florida}
\email{}

\maketitle

\begin{abstract}
With multiple crowd gatherings of millions of people every year in events ranging from pilgrimages to protests, concerts to marathons, and festivals to funerals; visual crowd analysis is emerging as a new frontier in computer vision. In particular, counting in highly dense crowds is a challenging problem with far-reaching applicability in crowd safety and management, as well as gauging political significance of protests and demonstrations. In this paper, we propose a novel approach that simultaneously solves the problems of counting, density map estimation and localization of people in a given dense crowd image. Our formulation is based on an important observation that the three problems are inherently related to each other making the loss function for optimizing a deep CNN decomposable. Since localization requires high-quality images and annotations, we introduce UCF-QNRF dataset that overcomes the shortcomings of previous datasets, and contains 1.25 million humans manually marked with dot annotations. Finally, we present evaluation measures and comparison with recent deep CNN networks, including those developed specifically for crowd counting. Our approach significantly outperforms state-of-the-art on the new dataset, which is the most challenging dataset with the largest number of crowd annotations in the most diverse set of scenes.

\keywords{Crowd Counting \and Localization \and Convolution Neural Networks \and Composition Loss}
\end{abstract}

\section{Introduction}

Counting dense crowds is significant both from socio-political and safety perspective. At one end of the spectrum, there are large ritual gatherings such as during pilgrimages that typically have large crowds occurring in known and pre-defined locations. Although they generally have passive crowds coming together for peaceful purposes, disasters have known to occur, for instance, during Love Parade \cite{helbing_epjds_2012} and Hajj \cite{guardian_hajj_2006}. For active crowds, such as expressive mobs in demonstrations and protests, counting is important both from political and safety standpoint. It is very common for different sides to claim divergent numbers for crowd gathering, inclined towards their political standing on the concerned issue. Beyond subjectivity and preference for certain political or social outcomes, the disparate counting estimates from opposing parties have a basis in numerical cognition as well. In humans, the results on subitizing \cite{piazza2002subitizing} suggest that once the number of observed objects increases beyond four, the brain switches from the exact Parallel Individuation System (PIS) to the inaccurate but scalable Approximate Number System (ANS) to count objects \cite{hyde2011two}. Thus, computer vision based crowd counting offers alternative fast and objective estimation of the number of people in such events. Furthermore, crowd counting is extendable to other domains, for instance, counting cells or bacteria from microscopic images \cite{zisserman-nips10,sirinukunwattana2016locality}, animal crowd estimates in wildlife sanctuaries \cite{arteta2016counting}, or estimating the number of vehicles at transportation hubs or traffic jams \cite{onoro2016towards}.

\begin{figure}[t]
\centering
\includegraphics[width=1\linewidth]{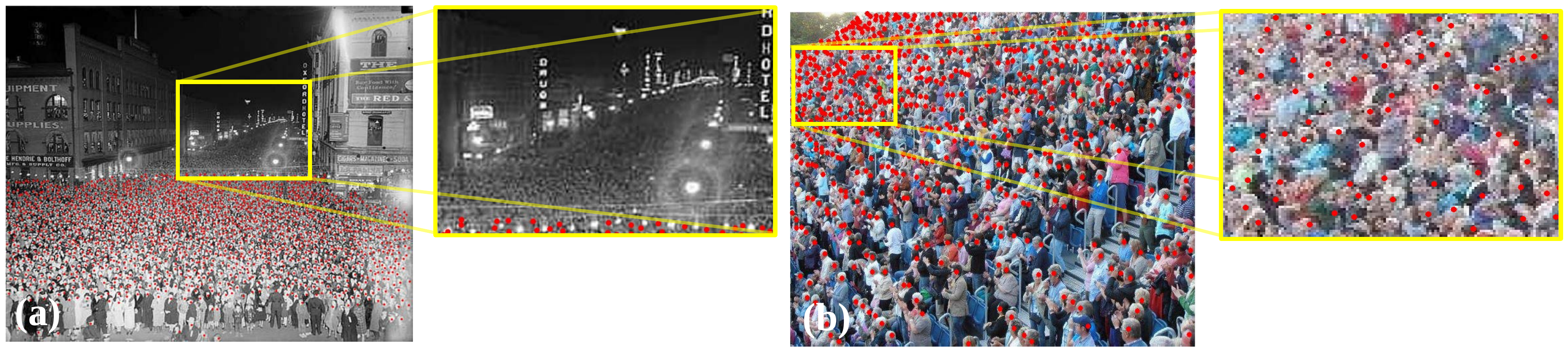}
\caption{{This figure highlights the problems due to low resolution images from two existing dense crowd datasets: (a) shows a case where the annotations were not done on parts of the images as it is virtually impossible to distinguish heads of neighboring people, while (b) shows a case where some of the locations / counts are erroneous and therefore not suitable for localization. The UCF-QNRF dataset proposed in this paper overcomes such issues.
}}
\label{figDatasetsProblems}
\end{figure}

In this paper, we propose a novel approach to crowd counting, density map estimation and localization of people in a given crowd image. Our approach stems from the observation that these three problems are very interrelated - in fact, they can be decomposed with respect to each other. Counting provides an estimate of the number of people / objects without any information about their location. Density maps, which can be computed at multiple levels, provide weak information about location of each person. Localization does provide accurate location information, nevertheless, it is extremely difficult to estimate directly due to its very sparse nature. Therefore, we propose to estimate all three tasks simultaneously, while employing the fact that each is special case of another one. Density maps can be `sharpened' till they approximate the localization map, whose integral should equal to the true count.

Furthermore, we introduce a new and the largest dataset to-date for training and evaluating \textbf{dense} crowd counting, density map estimation and localization methods, particularly suitable for training very deep Convolutional Neural Networks (CNNs). Though counting has traditionally been considered the primary focus of research, density map estimation and localization have significance and utility beyond counting. In particular, two applications are noteworthy: initialization / detection of people for tracking in dense crowds \cite{idrees2014tracking}; and rectifying counting errors from an automated computer vision algorithm. That is, a real user or analyst who desires to estimate the exact count for a real image \textit{without any error}, the results of counting alone are insufficient. The single number for an entire image makes it difficult to assess the error or the source of the error. However, the localization can provide an initial set of dot locations of the individuals, the user then can quickly go through the image and remove the false positives and add the false negatives. The count using such an approach will be much more accurate and the user can get $100\%$ precise count for the query image. This is particularly important when the number of image samples are few, and reliable counts are desired.

Prior to 2013, much of the work in crowd counting focused on low-density scenarios. For instance, UCSD dataset \cite{vasconcelos-cvpr08} contains $2,000$ video frames with $49,885$ annotated persons. The dataset is low density and low resolution compared to many recent datasets, where train and test splits belong to a single scene. WorldExpo'10 dataset \cite{zhang2015cross}, contains $108$ low-to-medium density scenes and overcomes the issue of diversity to some extent. UCF dataset \cite{idrees2013multi} contains 50 different images with counts ranging between $96$ and $4,633$ per image. Each image has a different resolution, camera angle, and crowd density. Although it was the first dataset for dense crowd images, it has problems with annotations (Figure \ref{figDatasetsProblems}) due to limited availability of high-resolution crowd images at the time. The ShanghaiTech crowd dataset \cite{zhang2016single} contains $1,198$ annotated images with a total of $330,165$ annotations. This dataset is divided into two parts: Part A contains $482$ images and Part B with $716$ images. The number of training images are $300$ and $400$ in both parts, respectively. Only the images in Part A contain high-density crowds, with $482$ images and $250$K annotations.

\renewcommand{\multirowsetup}{\centering}
\begin{table*}[t]
\centering
 \scalebox{1.35}{
 \tiny
 \setlength{\tabcolsep}{1pt}
 \begin{tabular}{c||c|c|c|c|c|c}
 \specialrule{1pt}{1pt}{1pt}
\hline
    \multirow{2}{2.0cm}{\bf{Dataset}} & \multirow{2}{1.0cm}{\bf{Number \\ Images}} & \multirow{2}{1.0cm}{\bf{Number \\Annotations}} & \multirow{2}{1.0cm}{\bf{Average\\Count}} & \multirow{2}{1.0cm}{\bf{Maximum\\Count}} & \multirow{2}{1.0cm}{\bf{Average\\Resolution}} & \multirow{2}{1.0cm}{\bf{Average\\Density}}\\
    &&&&&&\\
    \hline

 \hline
 UCF\_CC\_50 \cite{idrees2013multi} & 50 & 63,974 & 1279 & 4633 & 2101 $\times$ 2888 & 2.02 $\times 10^{-4}$ \\
 \hline
 WorldExpo'10 \cite{zhang2015cross} & 3980 & 225,216 & 56 & 334 & 576 $\times$ 720 & 1.36 $\times 10^{-4}$ \\
 \hline
 ShanghaiTech\_A \cite{zhang2016single} & 482 & 241,677 & 501 & 3139 & 589 $\times$ 868 & 9.33 $\times 10^{-4}$ \\
\hline
\hline
 \textbf{UCF-QNRF} & \textbf{1535} & \textbf{1,251,642} & \textbf{815} & \textbf{12865} & \textbf{2013 $\times$ 2902} & \textbf{1.12} $\mathbf{\times 10^{-4}}$ \\
 \hline
\specialrule{1pt}{1pt}{1pt}
\end{tabular}
}
\caption{{Summary of statistics of different datasets. UCF\_CC\_50 (44MB); WorldExpo'10 (325MB); ShanghaiTech\_A (67MB); and the proposed UCF-QNRF Dataset (4.33GB).}}
\label{tabDatasets}
\end{table*}

\begin{figure}[t]
\centering
\includegraphics[width=1\linewidth]{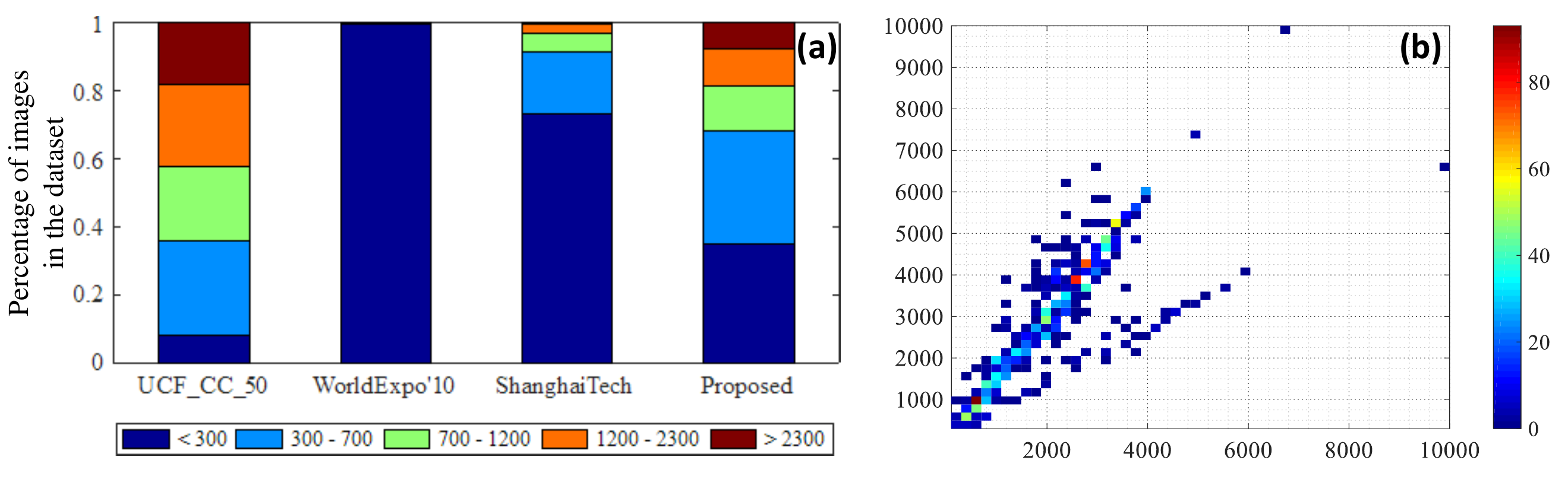}
\caption{{(a) This graph shows the relative distribution of image counts among the four datasets. The proposed UCF-QNRF dataset has a fair number of images from all five count ranges. (b) This graph shows a 2D histogram of image resolution for all the images in the new dataset. The x-axis shows the number of rows, while y-axis is the number of columns. Each bin ($500 \times 500$ pixels) is color-coded with the number of images that have the corresponding resolution.}}
\label{figDatasetStats}
\end{figure}

Table \ref{tabDatasets} summarizes the statistics of the multi-scene datasets for dense crowd counting. The proposed UCF-QNRF dataset has the most number of high-count crowd images and annotations, and a wider variety of scenes containing the most diverse set of viewpoints, densities and lighting variations. The resolution is large compared to WorldExpo'10 \cite{zhang2015cross} and ShanghaiTech \cite{zhang2016single}, as can be seen in Fig. \ref{figDatasetStats}(b). 
The average density, i.e., the number of people per pixel over all images is also the lowest, signifying high-quality large images. Lower per-pixel density is partly due to inclusion of background regions, where there are many high-density regions as well as zero-density regions. Part A of Shanghai dataset has high-count crowd images as well, however, they are severely cropped to contain crowds only. On the other hand, the new UCF-QNRF dataset contains buildings, vegetation, sky and roads as they are present in realistic scenarios captured in the wild. This makes this dataset more realistic as well as difficult. Similarly, Figure \ref{figDatasetStats}(a) shows the diversity in counts among the datasets. The distribution of proposed dataset is similar to UCF\_CC\_50 \cite{idrees2013multi}, however, the new dataset is $30$ and $20$ times larger in terms of number of images and annotations, respectively, compared to UCF\_CC\_50 \cite{idrees2013multi}. We hope the new dataset will significantly increase research activity in visual crowd analysis and will pave way for building deployable practical counting and localization systems for dense crowds.

The rest of the paper is organized as follows. In Sec. \ref{secRelatedWork} we review related work, and present the proposed approach for simultaneous crowd counting, density map estimation and localization in Sec. \ref{secProposedApproach}. The process for collection and annotation of the UCF-QNRF dataset is covered in Sec. \ref{secDataset}, while the three tasks and evaluation measures are motivated in Sec. \ref{secTasks}. The experimental evaluation and comparison are presented in Sec. \ref{secExperiments}. We conclude with suggestions for future work in Sec. \ref{secConclusion}.

\section{Related Work}\label{secRelatedWork}
Crowd counting is active an area of research with works tackling the three aspects of the problem: counting-by-regression \cite{ryan2009crowd}, \cite{zisserman-nips10}, \cite{idrees2013multi}, \cite{vasconcelos-cvpr08}, \cite{wang2015deep}, density map estimation \cite{zisserman-nips10}, \cite{fiaschi2012learning}, \cite{zhang2015cross}, \cite{pham2015count}, \cite{zhang2016single} and localization \cite{ma2015small}, \cite{rodriguez-iccv11b}.

Earlier regression-based approaches mapped global image features or a combination of local patch features to obtain counts \cite{kong2006viewpoint}, \cite{ccloy-shaogang-bmvc2012}, \cite{idrees2013multi}, \cite{chen2013cumulative}. Since these methods only produce counts, they cannot be used for density map estimation or localization. The features were hand-crafted and in some cases multiple features were used \cite{vasconcelos-cvpr08}, \cite{idrees2013multi} to handle low resolution, perspective distortion and severe occlusion. On the other hand, CNNs inherently learn multiple feature maps automatically, and therefore are now being extensively used for crowd counting and density map estimation.

CNN based approaches for crowd counting include \cite{lebanoff2015counting}, \cite{zhang2015cross}, \cite{zhang2016single}, \cite{onoro2016towards}, \cite{arteta2016counting}. Zhang \etal~\cite{zhang2015cross} train a CNN alternatively to predict density map and count in a patch, and then average the density map for all the overlapping patches to obtain density map for the entire image. Lebanoff~and~Idrees~\cite{lebanoff2015counting} introduce a normalized variant of the Euclidean loss function in a deep network to achieve consistent counting performance across all densities. The authors in \cite{zhang2016single} use three column CNN, each with different filter sizes to capture responses at different scales. The count for the image is obtained by summing over the predicted density map. Sindagi~and~Patel~\cite{sindagi2017generating} presented a CNN-based approach that incorporates global and local contextual information in an image to generate density maps. The global and local contexts are obtained by learning to classify the input image patches into various density levels, later fused with the output of a multi-column CNN to obtain the final density map. Similarly, in the approach by Sam~\etal~\cite{sam2017switching}, image patches are relayed to the appropriate CNN using a switching mechanism learnt during training. The independent CNN regressors are designed to have different receptive fields while the switch classifier is trained to relay the crowd scene patch to the best CNN regressor.

For localization in crowded scenes, Rodriguez~\etal~\cite{rodriguez-iccv11b} use density map as a regularizer during the detection. They optimize an objective function that prefers density map generated on detected locations to be similar to predicted density map \cite{zisserman-nips10}. This results in both better precision and recall. The density map is generated by placing a Gaussian kernel at the location of each detection. Zheng~\etal~\cite{ma2015small} first obtain density map using sliding window over the image through \cite{zisserman-nips10}, and then use integer programming to localize objects on the density maps. Similarly, in the domain of medical imaging, Sirinukunwattana~\etal~\cite{sirinukunwattana2016locality} introduced spatially-constrained CNNs for detection and classification of cancer nuclei. In this paper, we present results and analysis for simultaneous crowd counting, density map estimation, and localization using Composition Loss on the proposed UCF-QNRF dataset.

\begin{figure}[t]
\centering
\includegraphics[width=1\linewidth]{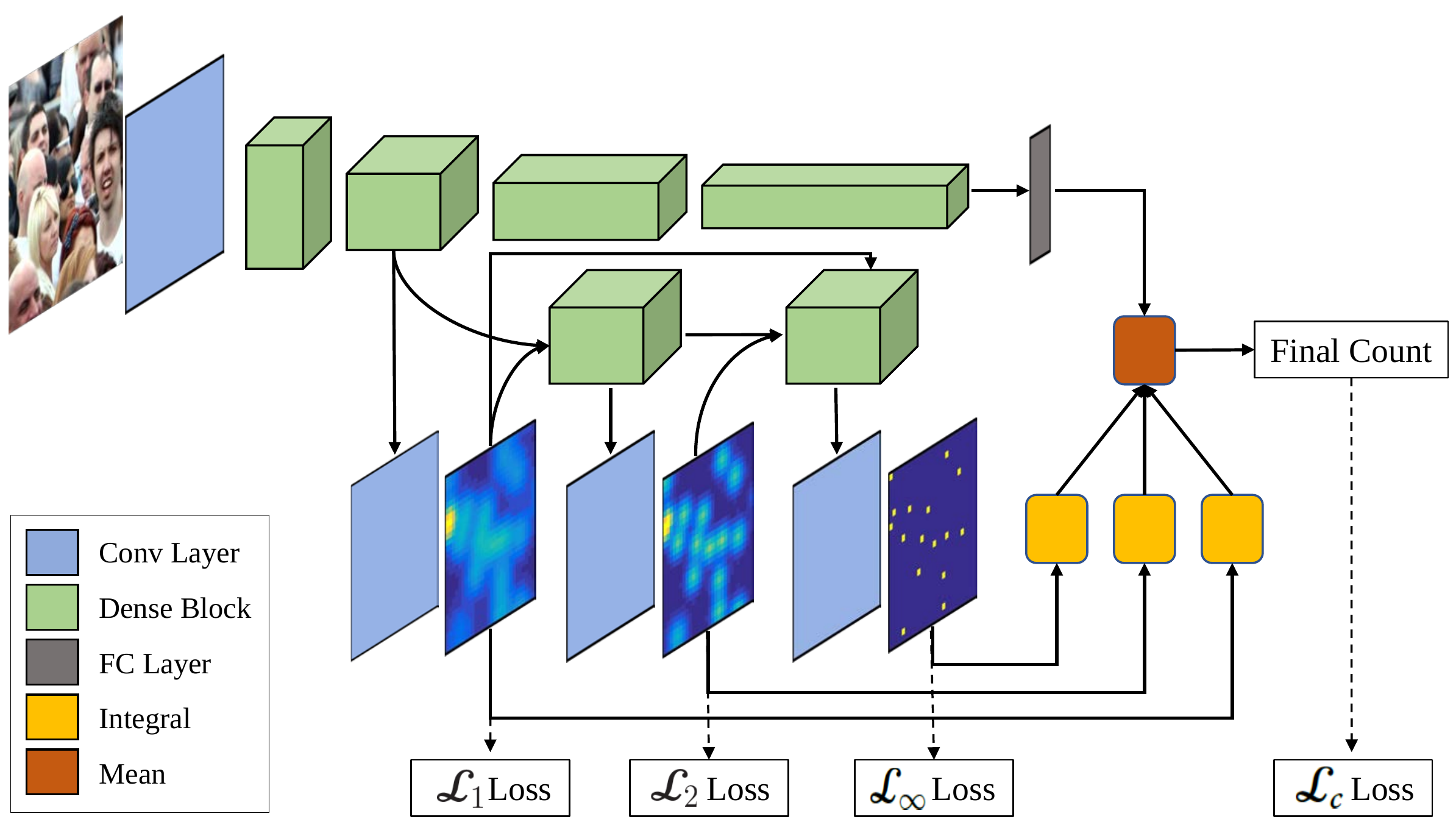}
\caption{{The figure shows the proposed architecture for estimating count, density and localization maps simultaneously for a given patch in an image. At the top is the base DenseNet which regresses only the counts. The proposed Composition Loss is implemented through multiple dense blocks after branching off the base network. We also test the effect of additional constraint on the density and localization maps (shown with amber and orange blocks) such that the count after integral in each should also be consistent with the groundtruth count. }}
\label{figModel}
\end{figure}

\section{Deep CNN with Composition Loss}\label{secProposedApproach}
In this section, we present the motivation for decomposing the loss of three interrelated problems of counting, density map estimation and localization, followed by details about the deep Convolutional Neural Network which can enable training and estimation of the three tasks simultaneously.

\subsection{Composition Loss}
Let $\mathbf{x} = [x,y]$ denote a pixel location in a given image, and $N$ be the number of people annotated with $\{\mathbf{x}_i : i = 1,2,\ldots N\}$ as their respective locations. Dense crowds typically depict heads of people as they are the only parts least occluded and mostly visible. In localization maps, only a single pixel is activated, i.e., set to $1$ per head, while all other pixels are set to $0$. This makes localization maps extremely sparse and therefore difficult to train and estimate. We observe that successive computation of `sharper' density maps which are relatively easier to train can aid in localization as well. Moreover, all three tasks should influence count, which is the integral over density or localization map. We use the Gaussian Kernel and adapt it for our problem of simultaneous solution for the three tasks.

Due to perspective effect and possibly variable density of the crowd, a single value of bandwidth, $\sigma$, cannot be used for the Gaussian kernel, as it might lead to well-defined separation between people close to the camera or in regions of low density, while excess blurring in other regions. Many images of dense crowds depict crowds in their entirety, making automatic perspective rectification difficult. Thus, we propose to define $\sigma_i$ for each person $i$ as the minimum of the $\ell_2$ distance to its nearest neighbor in spatial domain of the image or some maximum threshold, $\tau$. This ensures that the location information of each person is preserved precisely irrespective of default kernel bandwidth, $\tau$. Thus, the adaptive Gaussian kernel is given by,
\begin{equation}
D(\mathbf{x,\; \textit{f}(\cdot)}) = \sum_{i=1}^{N} \frac{1}{{ \sqrt {2\pi } \textit{f}(\sigma_i) }} \mathrm{exp}\bigg({{{ - \frac{ (x - x_i)^2 + (y - y_i)^2 } {2 \textit{f}(\sigma_i) ^2 }}}}\bigg),
\end{equation}
where the function $\textit{f}$ is used to produce a successive set of `sharper' density maps. We define $\textit{f}_k(\sigma)=\sigma^{1/k}$. Thus, $D_k = D(\mathbf{x},\; \textit{f}_k(\cdot))$. As can be seen when $k=1$, $D_k$ is a very smoothed-out density map using nearest-neighbor dependent bandwidth and $\tau$, whereas as $k\longrightarrow \infty$, $D_k$ approaches the binary localization map with a Dirac Delta function placed at each annotated pixel. Since each pixel has a unit area, the localization map assumes a unit value at the annotated location. For our experiments we used three density levels with last one being the localization map. It is also interesting to note that the various connections between density levels and base CNN also serve to provide intermediate supervision which aid in training the filters of base CNN towards counting and density estimation early on in the network.

Hypothetically, since integral over each estimated $\hat{D}_k$ yields a count for that density level, the final count can be obtained by taking the mean of counts from the density and localization maps as well as regression output from base CNN. This has two potential advantages: 1) the final count relies on multiple sources - each capturing count at a different scale. 2) During training the mean of four counts should equal the true count, which implicitly enforces an additional constraint that $\hat{D}_k$ should not only capture the density and localization information, but that each of their counts should also sum to the groundtruth count. For training, the loss function of density and localization maps is the mean square error between the predicted and ground truth maps, i.e. $\mathbcal{L}_k = \textrm{MSE}(\hat{D}_k, D_k)$, where $k=1,2,\;\textrm{and}\; \infty$, and regression loss, $\mathbcal{L}_c$, is Euclidean loss between predicted and groundtruth counts, while the final loss is defined as the weighted mean all four losses.

%
%
\begin{wraptable}{r}{7cm}
\vspace{-0.2in}
\centering
 \footnotesize
 \setlength{\textfloatsep}{5pt plus 1.0pt minus 2.0pt}
 \setlength{\floatsep}{5pt plus 1.0pt minus 2.0pt}
 \begin{tabular}{c|c|c}
 \specialrule{1.5pt}{1pt}{1pt}
 \hline
 \bf{Layer} & \bf{Output Size} & \bf{Filters} \\ [0.3ex]
 \hline
 \hline
 & 512 $\times$ \cross{28} & \\ \hline
 Density Level 1 & 1 $\times$ \cross{28} & \cross{1} conv \\ \hline
 \multirow{2}{*} {Density Level 2}  &  641 $\times$ \cross{28}   & \multicolumn{1}{l}{\conv{4}} \\ \cline{2-3}
 &1 $\times$ \cross{28}   & \multicolumn{1}{c}{\cross{1} conv} \\ \hline
 \multirow{2}{*} {Density Level $\infty$}  &  771 $\times$ \cross{28}   & \multicolumn{1}{l}{\conv{4}} \\ \cline{2-3}
 &1 $\times$ \cross{28}   & \multicolumn{1}{c}{\cross{1} conv} \\
 \hline
\specialrule{1.5pt}{1pt}{1pt}
\end{tabular}
\caption{{This table shows the filter dimensions and output of the three density layer blocks appended to the network in Fig. \ref{figModel}.}}
\label{tabDensityModule}
\vspace{-0.1in}
\end{wraptable}

\subsection{DenseNet with Composition Loss}

We use DenseNet~\cite{huang2016densely} as our base network. It consists of $4$ Dense blocks where each block has a number of consecutive $1 \times 1$ and $3 \times 3$ convolutional layers. Each dense block (except for the last one) is followed by a Transition layer, which reduces the number of feature-maps by applying $1 \times 1$ convolutions followed by $2 \times 2$ average pooling with stride 2. In our experiments we used DenseNet-201 architecture. It has $\{6, 12, 48, 32\}$ sets of $1 \times 1$ and $3 \times 3$ convolutional layers in the four dense blocks, respectively.

For density map estimation and localization, we branch out from DenseBlock2 and feed it to our Density Network (see Table \ref{tabDensityModule}). The density network introduces $2$ new dense blocks and three $1 \times 1$  convolutional layers. Each dense block has features computed at the previous step, concatenated with all the density levels predicted thus far as input, and learns features aimed at computing the current density / localization map. We used $1 \times 1$ convolutions to get the output density map from these features. Density Level $1$ is computed directly from DenseBlock2 features.

We used Adam solver with a step learning rate in all our experiments. We used $0.001$ as initial learning rate and reduce the learning rate by a factor of $2$ after every $20$ epochs. We trained the entire network for $70$ epoch with a batch size of $16$.

\section{The UCF-QNRF Dataset}\label{secDataset}
\smallskip
\noindent\textbf{Dataset Collection.} The images for the dataset were collected from three sources: Flickr, Web Search and the Hajj footage. The Hajj images were carefully selected so that there are multiple images that capture different locations, viewpoints, perspective effects and times of the day. For Flickr and Web Search, we manually generated the following queries: \textsc{crowd}, \textsc{hajj}, \textsc{spectator crowd}, \textsc{pilgrimage}, \textsc{protest crowd} and \textsc{concert crowd}. These queries were then passed onto the Flickr and Google Image Search APIs. We selected desired number of images for each query to be $2000$ for Flickr and $200$ for Google Image Search. The search sorted all the results by \textsc{relevance} incorporating both titles and tags, and for Flickr we also ensured that only those images were downloaded for which original resolutions were permitted to be downloaded (through the \textsc{url\_o} specifier). The static links to all the images were extracted and saved for all the query terms, which were then downloaded using the respective APIs. The images were also checked for duplicates by computing image similarities followed by manual verification and discarding of duplicates.
\vskip 0.1in

\noindent\textbf{Initial Pruning.} The initial set of images were then manually checked for desirability. Many of the images were pruned due to one or more of the following reasons:
\begin{itemize}[noitemsep]
  \item Scenes that did not depict crowds at all or low-density crowds
  \item Objects or visualizations of objects other than humans
  \item Motion blur or low resolution
  \item Very high perspective effect that is camera height is similar to average human height
  \item Images with watermarks or those where text occupied more than 10\% of the image
\end{itemize}

In high-density crowd images, it is mostly the heads that are visible. However, people who appear far away from the camera become indistinguishable beyond a certain distance, which depends on crowd density, lighting as well as resolution of the camera sensor. During pruning, we kept those images where the heads were separable visually. Such images were annotated with the others, however, they were cropped afterwards to ensure that regions with problematic annotations or those with none at all due to difficulty in recognizing human heads were discarded.

We performed the entire annotation process in two stages. In the first stage, un-annotated images were given to the \textit{annotators}, while in the second stage, the images were given to \textit{verifiers} who corrected any mistakes or errors in annotations. There were $14$ annotators and $4$ verifiers, who clocked ~$1,300$ and ~$200$ hours respectively. In total, the entire procedure involved ~$2,000$ human-hours spent through to its completion.
\vskip 0.1in

\noindent\textbf{Statistics.} The dataset has $1,535$ jpeg images with $1,251,642$ annotations. The train and test sets were created by sorting the images with respect to absolute counts, and selecting every 5th image into the test set. Thus, the training and test set consist of $1201$ and $334$ images, respectively. The distribution of images from [Flickr, Web, Hajj] for the train and test are [1078, 84, 39] and [306, 21, 7], respectively. In the dataset, the minimum and maximum counts are $49$ and $12,865$, respectively, whereas the median and mean counts are $425$ and $815.4$, respectively.

\section{Definition and Quantification of Tasks}\label{secTasks}
In this section, we define the three tasks and the associated quantification measures.

\noindent\textbf{Counting:} The first task involves estimation of count for a crowd image $i$, given by $\mathbf{c}_i$. 
Although this measure does not give any information about location or distribution of people in the image, this is still very useful for many applications, for instance, estimating size of an entire crowd spanning several square kilometers or miles. For the application of counting large crowds, Jacob's Method \cite{jacobmethod67} due to Herbert Jacob is typically employed which involves dividing the area $\mathbf{A}$ into smaller sections, finding the average number of people or density $\mathbf{d}$ in each section, computing the mean density $\bar{\mathbf{d}}$ and extrapolating the results to entire region. 
However, with automated crowd counting, it is now possible to obtain counts and density for multiple images at different locations, thereby, permitting the more accurate integration of density over entire area covered by crowd. Moreover, counting through multiple aerial images requires cartographic tools to map the images onto the earth to compute ground areas. The density here is defined as the number of people in the image divided by ground area covered by the image. 
We propose to use the same evaluation measures as used in literature for this task: the Mean Absolute Error (C-MAE), Mean Squared Error (C-MSE) with the addition of Normalized Absolute Error (C-NAE).

\smallskip
\noindent\textbf{Density Map Estimation} amounts to computing per-pixel density at each location in the image, thus preserving spatial information about distribution of people. This is particularly relevant for safety and surveillance, since very high density at a particular location in the scene can be catastrophic \cite{guardian_hajj_2006}. This is different from counting since an image can have counts within safe limits, while containing regions that have very high density. This can happen due to the presence of empty regions in the image, such as walls and sky for mounted cameras; and roads, vehicles, buildings and forestation in aerial cameras. 
The metrics for evaluating density map estimation are similar to counting, except that they are per-pixel, i.e., the per-pixel Mean Absolute Error (DM-MAE) and Mean Squared Error (DM-MSE). 
Finally, we also propose to compute the 2D Histogram Intersection (DM-HI) distance after normalizing both the groundtruth and estimated density maps. This discards the effect of absolute counts and emphasizes the error in distribution of density compared to the groundtruth. 

\smallskip
\noindent\textbf{Localization:} The ideal approach to crowd counting would be to detect all the people in an image and then count the number of detections. But since dense crowd images contain severe occlusions among individuals and fewer pixels per person for those away from the camera, this is not a feasible solution. This is why, most approaches to crowd counting bypass explicit detection and perform direct regression on input images. However, for many applications, the precise location of individuals is needed, for instance, to initialize a tracking algorithm in very high-density crowd videos.


To quantify the localization error, estimated locations are associated with the ground truth locations through 1-1 matching using greedy association, followed by computation of Precision 
and Recall 
at various distance thresholds ($1,2,3, \ldots, 100$ pixels). The overall performance of the localization task is then computed through area under the Precision-Recall curve, $\textrm{L-AUC}$.  

\section{Experiments}\label{secExperiments}

Next, we present the results of experiments for the three tasks defined in Section \ref{secTasks}.

\subsection{Counting}\label{subsecRegression}
\begin{figure}[t]
\centering
\includegraphics[width=1\linewidth]{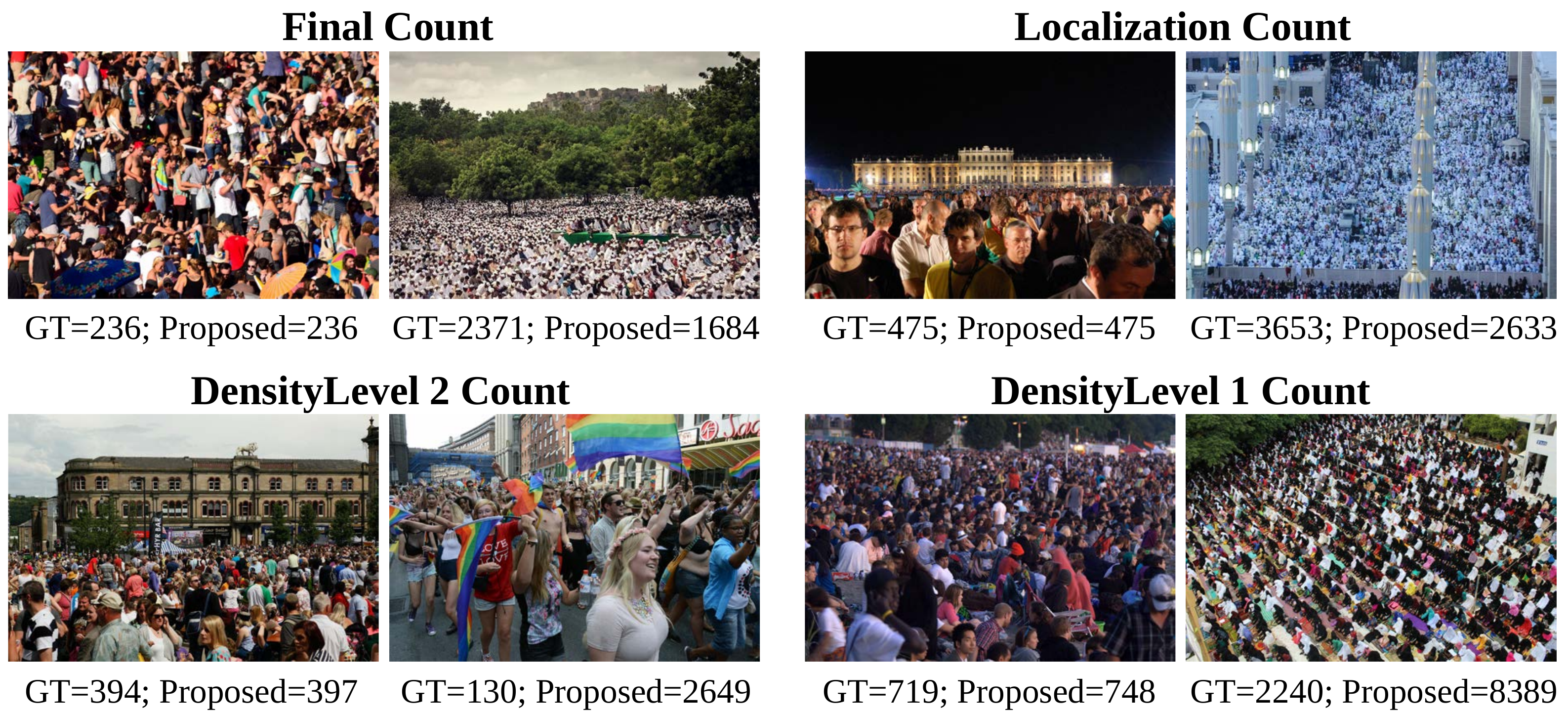}
\caption{{This figure shows pairs of images where the left image in the pair has the lowest counting error while the right image has the highest counting error with respect to the four components of the Composition Loss.}}
\label{fig:regResult}
\end{figure}

\begin{wraptable}{r}{7cm}
\vspace{-0.2in}
\centering
\small
 \begin{tabular}{ c || c | c| c }
 \specialrule{1.5pt}{1pt}{1pt}
 \hline
 Method & C-MAE & C-NAE & C-MSE \\ [0.5ex]
 \hline\hline
 Idrees \etal \cite{idrees2013multi}* & 315 & 0.63 & 508 \\
 \hline
  MCNN \cite{zhang2016single} & 277 & 0.55 & 426 \\
 \hline
 Encoder-Decoder \cite{badrinarayanan2015segnet} & 270 & 0.56 & 478 \\ 
 \hline
  CMTL \cite{sindagi2017cnn} & 252 & 0.54 & 514 \\
 \hline
 SwitchCNN \cite{sam2017switching} & 228 & 0.44  & 445  \\
 \hline
 Resnet101 \cite{he2016deep}* & 190 & 0.50 & 277 \\
 \hline
 Densenet201 \cite{huang2016densely}* & 163 & 0.40 & 226 \\ 
 \hline
 \hline
 \bf{Proposed} & \bf{132} & \bf{0.26} & \bf{191}\\
 \hline
 \specialrule{1.5pt}{1pt}{1pt}
\end{tabular}
\caption{{We show counting results obtained using  state-of-the-art methods in comparison with the proposed approach. Methods with `*' regress counts without computing density maps.}}
\label{table:CountingSummary}
\vspace{-0.1in}
\end{wraptable}
For counting, we evaluated the new UCF-QNRF dataset using the proposed method which estimates counts, density maps and locations of people simultaneously with several state-of-the-art deep neural networks \cite{badrinarayanan2015segnet}, \cite{he2016deep}, \cite{huang2016densely} as well as those specifically developed for crowd counting \cite{zhang2016single}, \cite{sindagi2017cnn}, \cite{sam2017switching}. To train the networks, we extracted patches of sizes $448, 224$ and $112$ pixels at random locations from each training image. While deciding on image locations to extract patch from, we assigned higher probability of selection to image regions with higher count. We used mean square error of counts as the loss function. At test time, we divide the image into a grid of $224 \times 224$ pixel cells - zero-padding the image for dimensions not divisible by 224 - and evaluate each cell using the trained network. Final image count is given by aggregating the counts in all cells. Table \ref{table:CountingSummary} summarizes the results which shows the proposed network significantly outperforms the competing deep CNNs and crowd counting approaches. In Figure \ref{fig:regResult}, we show the images with the lowest and highest error in the test set, for counts obtained through different components of the Composition Loss.

\subsection{Density Map Estimation}\label{subsecDensityEstimation}

\begin{wraptable}{r}{7cm}
\vspace{-0.2in}
\centering
\small
\centering
\small
 \begin{tabular}{ c || c | c | c }
 \specialrule{1.5pt}{1pt}{1pt}
 \hline
 Method & DM-MAE & DM-MSE & DM-HI \\
 \hline\hline
 MCNN \cite{zhang2016single} & 0.006670 & 0.0223 & 0.5354 \\
 \hline
 SwitchCNN \cite{sam2017switching} &0.005673 & 0.0263 & 0.5301 \\
 \hline
 CMTL \cite{sindagi2017cnn} &0.005932 &0.0244 & 0.5024  \\
 \hline
 \hline
 \bf{Proposed} & \textbf{0.00044} & \textbf{0.0017} & \textbf{0.9131} \\
 \hline
 \specialrule{1.5pt}{1pt}{1pt}
\end{tabular}
\caption{{Results for Density map estimation:  We show results on Histogram intersection (HI), obtained using existing state-of-the-art methods compared to the proposed approach.}}
\label{table:densityestimationSummary}
\vspace{-0.1in}
\end{wraptable}

For density map estimation, we describe and compare the proposed approach with several methods that directly regress crowd density during training. Among the deep learning methods, MCNN \cite{zhang2016single} consists of three columns of convolution networks with different filter sizes to capture different head sizes and combines the output of all the columns to make a final density estimate. SwitchCNN \cite{sam2017switching} uses a similar three column network; however, it also employs a switching network that decides which column should exclusively handle the input patch. CMTL \cite{sindagi2017cnn} employs a multi-task network that computes a high level prior over the image patch (crowd count classification) and density estimation. These networks are specifically designed for crowd density estimation and their results are reported in first three rows of Table \ref{table:densityestimationSummary}. The results of proposed approach are shown in the bottom row of Table \ref{table:densityestimationSummary}. The proposed approach outperforms existing approaches by an order of magnitude.

\subsection{Localization}\label{subsecLocalization}

For the localization task, we adopt the same network configurations used for density map estimation to perform localization. To get the accurate head locations, we post-process the outputs by finding the local peaks / maximums based on a threshold, also known as non-maximal suppression. Once the peaks are found, we match the predicted location with the ground truth location using 1-1 matching, and compute precision and recall. We use different distance thresholds as the pixel distance, i.e., if the detection is within the a particular distance threshold of the groundtruth, it is treated as True Positive, otherwise it is a False Positive. Similarly, if there is no detection within a groundtruth location, it becomes a False Negative.

\begin{figure}[t]
\centering
\includegraphics[width=1\linewidth]{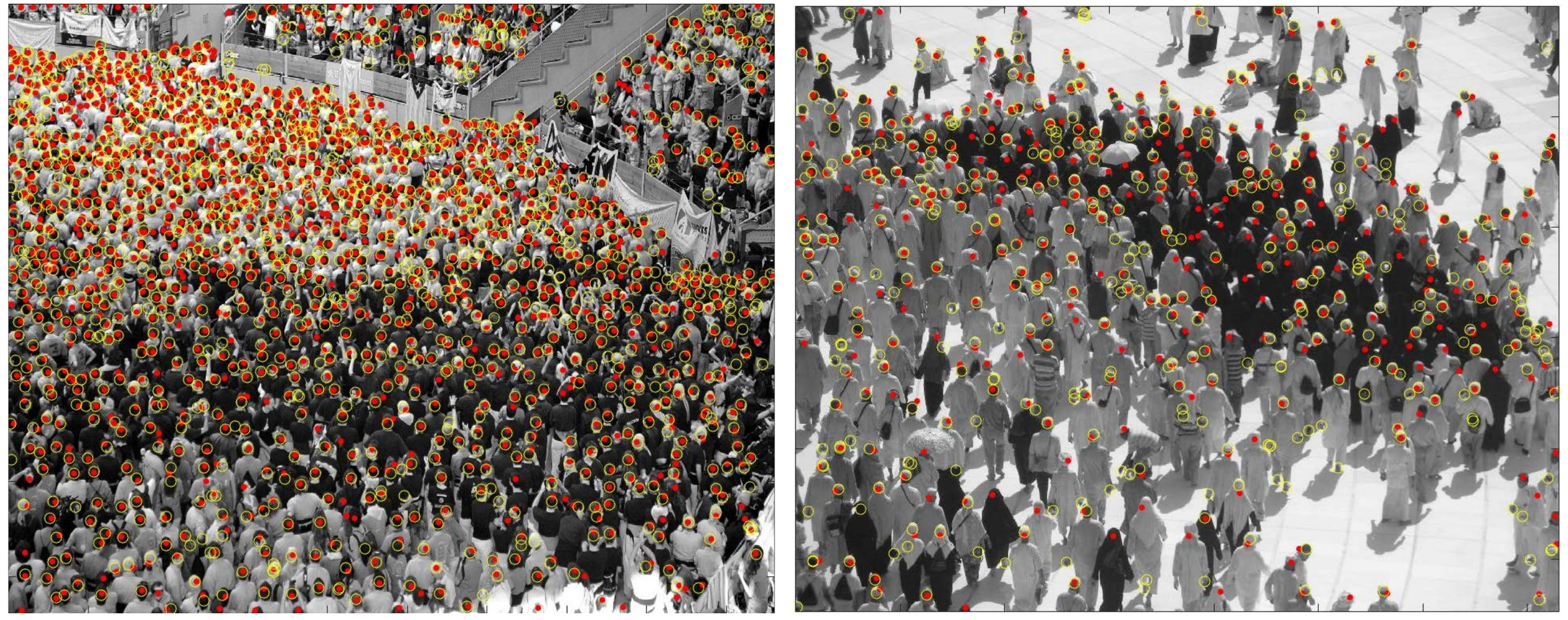}
\caption{{Two examples of localization using the proposed approach. Ground truth is depicted in red and predicted locations after threshold are shown in yellow.}}
\label{fig:Local}
\end{figure}

The results of localization are reported in Table \ref{table:localizationSummary}. This table shows that DenseNet \cite{huang2016densely} and Encoder-Decoder \cite{badrinarayanan2015segnet} outperform ResNet \cite{he2016deep} and MCNN \cite{zhang2016single}, while the proposed approach is superior to all the compared methods. The performance on the localization task is dependent on post-processing, which can alter results. Therefore, finding optimal strategy for localization from neural network output or incorporating the post-processing into the network is an important direction for future research. 
We also show some qualitative results of localization in Figure \ref{fig:Local}. The red dots represent the groundtruth while yellow circles are the locations estimated by the our approach.

\begin{table}[t]
\centering
{\renewcommand{\arraystretch}{1}
\begin{tabular}{c || c | c | c  }
\specialrule{1.5pt}{1pt}{1pt}
\hline
Method & Av.  Precision & Av. Recall  & L-AUC \\
\hline\hline
MCNN \cite{zhang2016single} & 59.93\% & 63.50\% & 0.591 \\
\hline
ResNet74 \cite{he2016deep} & 61.60\% & 66.90\%  & 0.612 \\
\hline
DenseNet63 \cite{huang2016densely} & 70.19\% & 58.10\% & 0.637 \\ 
\hline
Encoder-Decoder \cite{badrinarayanan2015segnet} & 71.80\% & 62.98\%  & 0.670\\ 
\hline
\textbf{Proposed} & \textbf{75.8\%}  & \textbf{59.75\%}  &  \textbf{0.714} \\
\hline
\specialrule{1.5pt}{1pt}{1pt}
\end{tabular}
}
\caption{{This table shows the localization results averaged over four distance thresholds for different methods. We show Average Precision, Average Recall and AUC metrics.}}
\label{table:localizationSummary}
\end{table}

\subsection{Ablation Study}

\begin{table}[t]
\centering
\begin{tabular}{l|ccc|ccc|ccc|ccc}
 \specialrule{1.5pt}{1pt}{1pt}
 \hline
\multirow{2}{*}
{\bf{Experiment}}   &       & \textbf{Count}  &       &       & $D_\infty$   &       &       & $D_2$     &       &       & $D_1$     &        \\
               & MAE   & MSE    & NAE   & MAE   & MSE    & NAE   & MAE   & MSE    & NAE   & MAE   & MSE    & NAE    \\
\hline
BaseNetwork   & 163 & 227 & 0.395 & -   & -   & -     & -   & -   & -     & -   & -    & -     \\
\hline
DenseBlock4 & 148 & 265 & 0.385  & 382 & 765 & 0.956 & 879 & 1235 & 3.892 & 2015 & 4529  & 4.295  \\
DenseBlock3 & 144 & 236 & 0.363  & 295 & 687 & 0.721 & 805 & 1159 & 3.256 & 1273 & 2936  & 3.982 \\
\hline
$D_1$ only & 141 & 233 & 0.261  & - & - & - & - & - & - & 1706 & 2496  & 5.677 \\
$D_1$ \& $D_2$ only & 137 & 208 & 0.251  & - & - & - & 691 & 1058 & 2.459 & 1887 & 3541  & 6.850 \\
\hline
Concatenate   & 139 & 223 & 0.264 & 258 & 508 & 0.634 & 718 & 1096 & 3.570 & 1910 & 4983  & 6.574 \\
Mean          & 150 & 341 & 0.271 & 405 & 710 & 1.135  & 1015 & 2099 & 2.916 & 1151 & 3170  & 3.283 \\
\hline
Proposed    & 132 & 191 & 0.258 & 236 & 408 & 0.506 & 682 & 922 & 2.027  & 1629 & 3600  & 4.396 \\
\specialrule{1.5pt}{1pt}{1pt}
\end{tabular}
\caption{{This table shows the results of ablation study. $D_\infty$ corresponds to the results of counting using localization map estimation, while $D_2$ and $D_1$ represent results from the two density maps, respectively.}}
\label{table:Ablation}
\end{table}

We performed an ablation study to validate the efficacy of composition loss introduced in this paper, as well as various choices in designing the network. These results are shown in Table \ref{table:Ablation}. Next, we describe and provide details for the experiment corresponding to each row in the table.

\smallskip

\noindent\textbf{BaseNetwork:} This row shows the results with base network of our choice, which is DenseNet201. A fully-connected layer is appended to the last layer of the network followed by a single neuron which outputs the count. The input patch size is \cross{224}.

\smallskip

\noindent\textbf{DenseBlock4}: This experiment studies the effect of connecting the Density Network (Table \ref{tabDensityModule}) containing the different density levels with DenseBlock4 of the base DenseNet instead of DenseBlock2. Since DenseBlock4 outputs feature maps of size \cross{7}, we therefore used deconvolution layer with stride 4 to upsample the features before feeding in to our Density Network.

\smallskip

\noindent\textbf{DenseBlock3}: This experiment is similar to \textbf{DenseBlock4}, except that we connect our Density Network to Denseblock3 of the base network. DenseBlock3 outputs feature maps which are \cross{14} in spatial dimensions,  whereas we intend to predict density maps of spatial dimension \cross{28}, so we upsample the feature maps by using deconvolution layer before feeding them to the proposed Density Network.

\smallskip

\noindent\textbf{$D_1$ only}: This row represents the results if we use Density Level 1 only in the Density Network along with regression of counts in the base network. The results are much worse compared to the proposed method which uses multiple levels in the Composition Loss.

\smallskip

\noindent\textbf{$D_1$ and $D_2$ only}: Similar to $D_1$ \textbf{only}, this row represents the results if we use Density Levels 1 and 2 and do not use the $D_\infty$ in the Density Network. Incorporation of another density level improves results slightly in contrast to a single density level.

\smallskip

\noindent\textbf{Concatenate}: Here, we take the sum of the two density and one localization map to obtain $3$ counts. We then concatenate these counts to the output of fully-connected layer of the base network to predict count from the single neuron. Thus, we leave to the optimization algorithm to find appropriate weights for these 3 values along with the rest of 1920 features of the fully-connected layer.

\smallskip

\noindent\textbf{Mean}: We also tested the effect of using equal weights for counts obtained from the base network and three density levels. We take sum of each density / localization map and take the mean of 4 values (2 density map sums, one localization sum, and one count from base network). We treat this mean value as final count output - both during training and testing. Thus, this imposes the constraint that not only the density and localization map correctly predict the location of people, but also their counts should be consistent with groundtruth counts irrespective of predicted locations.

\smallskip

\noindent\textbf{Proposed}: In this experiment, the Density Network is connected with the DenseBlock2 of base network, however, the Density Network simply outputs two density and one localization maps, none of which are connected to count output (see Figure \ref{figModel}).

\smallskip

In summary, these results show that the Density Network contributes significantly to performance on the three tasks. It is better to branch out from the middle layers of the base network, nevertheless the idea of multiple connections back and forth from the base network and Density Network is an interesting direction for further research. Furthermore, enforcing counts from all sources to be equal to the groundtruth count slightly worsens the counting performance. Nevertheless, it does help in estimating better density and localization maps. Finally, the decrease in error rates from the right to left in Table \ref{table:Ablation} highlights the positive influence of the proposed Composition Loss.

\section{Conclusion}\label{secConclusion}
This paper introduced a novel method to estimate counts, density maps and localization in dense crowd images. We showed that these three problems are interrelated, and can be decomposed with respect to each other through Composition Loss which can then be used to train a neural network. We solved the three tasks simultaneously with the counting performance benefiting from the density map estimation and localization as well. We also proposed the large-scale UCF-QNRF dataset for dense crowds suitable for the three tasks described in the paper. We provided details of the process of dataset collection and annotation, where we ensured that only high-resolution images were curated for the dataset. 
Finally, we presented extensive set of experiments using several recent deep architectures, and show how the proposed approach is able to achieve good performance through detailed ablation study. We hope the new dataset will prove useful for this type of research, with applications in safety and surveillance, design and expansion of public infrastructures, and gauging political significance of various crowd events.

\smallskip
\noindent\textbf{Acknowledgment:} This work was made possible in part by NPRP grant number NPRP~7-1711-1-312 from the Qatar National Research Fund (a member of Qatar Foundation). The statements made herein are solely the responsibility of the authors.

\bibliographystyle{splncs04}
\bibliography{2324_bib}

\end{document}